\relax
\documentclass[letterpaper]{article}
\usepackage{aaai17_mod}
\usepackage{times}
\usepackage{helvet}
\usepackage{courier}
\usepackage[utf8]{inputenc} 
\usepackage[T1]{fontenc}    
\usepackage{url}            
\usepackage{booktabs}       
\usepackage{amsfonts}       
\usepackage{nicefrac}       
\usepackage{microtype}      

\usepackage{epsfig}
\usepackage{graphicx}
\usepackage{amsmath}
\usepackage{amssymb}
\usepackage{array}

\title{Learning Latent Sub-events in Activity Videos Using Temporal Attention Filters}
\author{AJ Piergiovanni\thanks{These authors contributed equally to the paper.}, Chenyou Fan\footnotemark[1], and Michael S. Ryoo\\
School of Informatics and Computing, Indiana University, Bloomington, IN 47408 \\
\texttt{\{ajpiergi,fan6,mryoo\}@indiana.edu}\\
}

\begin{document}

\maketitle

\begin{abstract}

In this paper, we newly introduce the concept of \emph{temporal attention filters}, and describe how they can be used for human activity recognition from videos. Many high-level activities are often composed of multiple temporal parts (e.g., sub-events) with different duration/speed, and our objective is to make the model explicitly learn such temporal structure using multiple attention filters and benefit from them. Our temporal filters are designed to be fully differentiable, allowing end-of-end training of the temporal filters together with the underlying frame-based or segment-based convolutional neural network architectures. This paper presents an approach of learning a set of optimal static temporal attention filters to be shared across different videos, and extends this approach to dynamically adjust attention filters per testing video using recurrent long short-term memory networks (LSTMs). This allows our temporal attention filters to learn latent sub-events specific to each activity. We experimentally confirm that the proposed concept of temporal attention filters benefits the activity recognition, and we visualize the learned latent sub-events.
\end{abstract}

\section{Introduction}

Human activity recognition is the problem of identifying events performed by humans given a video input. It is formulated as a binary (or multiclass) classification problem of outputting activity class labels, and researchers have been studying better features, representations, and learning algorithms to improve the classification \cite{aggarwal11}. Such classification not only allows categorization of videos pre-segmented to contain one single activity, but also enables the `detection' of activities from streaming videos together with temporal window proposal methods like the sliding window or selective search \cite{ryoo15hri}. Activity recognition is an important problem with many societal applications including smart surveillance, video search/retrieval, intelligent robots, and other monitoring systems.



Particularly, in the past 2-3 years, activity recognition approaches taking advantage of convolutional neural networks (CNNs) have received a great amount of attention. Motivated by the success of image-based object recognition using CNNs, researchers attempted developing CNNs for videos. Some approaches directly took advantage of image-based CNN architectures by applying them to every video frame \cite{jain14,google15}, while some tried to learn 3-D XYT spatio-temporal convolutional filters from short video segments \cite{c3d}. In order to represent each video, temporal pooling (e.g., max/average pooling) were often applied on top of multiple (sampled) per-frame or per-video-segment CNNs \cite{jain14,karpathy14,simonyan14,google15,c3d}. Similar to the object recognition, these approaches obtained superior results compared to traditional approaches of using hand-crafted features.

However, in terms of learning and considering activities' temporal structure in videos, previous CNN approaches were limited. Many high-level activities are often composed of multiple temporal parts (i.e., sub-events) with different duration/speed, but approaches to \emph{learn} explicit activity temporal structure together with CNN parameters have not been studied in depth. For instance, the typical strategy of taking max (or average) pooling over sampled per-frame or per-segment CNN responses \cite{jain14,simonyan14,google15,c3d} completely ignores such temporal structure in longer activity videos. \cite{ryoo15} showed a potential that making the system consider multiple video sub-intervals using a temporal pyramid benefits the recognition, but it was done with predetermined intervals without any learning. LSTM-based recurrent neural network approaches \cite{google15,yeong16} were able to process per-frame CNN responses sequentially, but no explicit sub-event or interval learning was attempted.

What we need instead is an approach that explicitly `learns' to focus on important sub-intervals of the activity videos while also optimizing their temporal resolution for the recognition. This is a challenging problem since we want to make this sub-event learning done in an end-to-end fashion together with the training of underlying CNN parameters. Furthermore, it is often the case that ground truth labels of sub-events to be learned are not provided, making them latent variables.


\begin{figure*}
\begin{center}
   \includegraphics[width=0.75\linewidth]{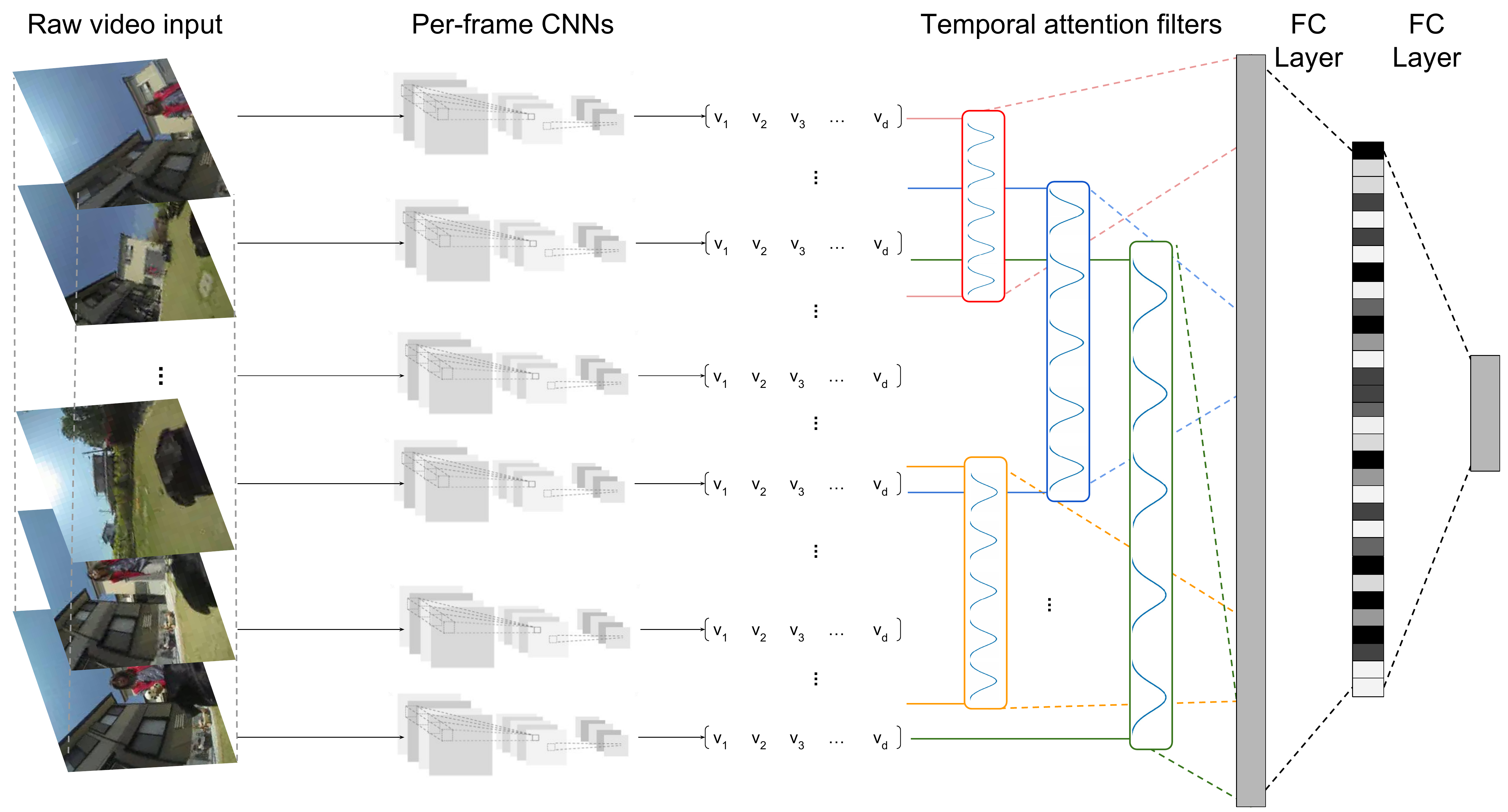}
\end{center}
   \caption{Illustration of our overall recognition architecture with temporal attention filters. $M$ number of temporal filters are learned to focus on different temporal part of video frame features (i.e., latent sub-events). Each filter is composed of a set of Gaussian filters which take a weighted sum of local information. Outputs of the temporal filters are concatenated, and attached with a fully connected layers to perform video classification. }
\label{fig:pipeline}		
\end{figure*}




This paper presents a new video classification approach that overcomes such limitations using temporal attention filters. We newly introduce the concept of fully differentiable \emph{temporal attention filters} and describe how they can be learned and used to enable better recognition of human activities from videos. The main idea is to make the network learn and take advantage of multiple temporal attention filters to be applied on top of per-frame CNNs (Figure \ref{fig:pipeline}). Each learned attention filter corresponds to a particular sub-interval of the activity the system should focus on (i.e., latent sub-events), and is represented with its center location, duration, and resolution in the relative temporal coordinate. Our approach abstracts per-frame (or per-segment) CNN responses within the sub-interval corresponding to the attention filter, allowing the system to use their results for the classification. Notably, our temporal attention filters are designed to be fully differentiable, motivated by the spatial attention filters for images \cite{attention15}. This allows the \emph{end-to-end} training of the parameters deciding attention filters; the system learns temporal attention filters jointly with the parameters of the underlying per-frame (or per-segment) CNNs. As a result, our temporal filters are trained to be optimized for the recognition tasks, automatically learning its location/scale from the training data without sub-event labels.

The paper not only presents an approach of learning optimal static temporal attention filters to be shared across different videos, but also present an approach of dynamically adjusting attention filters per testing video using recurrent long short-term memory cells (LSTMs). Instead of learning static temporal filters who location/duration/resolution is shared by all videos, our LSTM based approach dynamically and adaptively adjusts its filter parameters depending on the video, by going through multiple iterations. Our proposed approach is able to function in conjunction with any per-frame or per-video-segment CNNs as well as with other types of feature representations (e.g., Fisher vectors), making it very generally applicable for many video understanding scenarios.

\section{Previous works}

As described in the introduction, the direction of using convolutional neural networks for video classification is becoming increasingly popular, since it allows end-to-end training of convolutional filters optimized for the training data. \cite{simonyan14} used optical flows in addition to image feature. \cite{google15} tested multiple different types of pooling strategies on top of per-frame CNNs, and found that the simple global max pooling of per-frame features over the entire interval performs the best for sports videos. \cite{karpathy14} also tried multiple different (temporal) pooling strategies, gradually combining per-frame CNN responses over a short interval using their `slow fusion'. \cite{c3d} proposed to do XYT convolution, learning space-time convolutional filters. \cite{tdd15} used local CNN feature maps around tracklets in videos.

However, these prior works focused only on capturing dynamics in very short video intervals without much consideration on long-term temporal structure of activity videos. Optical flow only captures differences between two consecutive frames \cite{simonyan14}. Even with the video-based 3-D XYT CNNs \cite{c3d} or trajectory CNNs \cite{tdd15}, only the temporal dynamics within short intervals with a fixed duration (e.g., 15 frames) were captured without considering longer-term structure or attempting to learn latent sub-events. \cite{ryoo15} showed a potential that considering temporal structure in terms of sub-intervals (e.g., temporal pyramid) may benefit the recognition, but they did not attempt any learning. Similarly, \cite{li16} considered multiple different temporal scales, but learning of how the system should choose such scales were not attempted. \cite{varol16} also used fixed intervals. Recurrent neural networks such as LSTMs were also used to model sequences \cite{google15}, but they were unable to explicitly consider different temporal sub-intervals and their structure. That is, learning to consider different intervals with different temporal resolution was not possible, and no sub-event learning as involved. \cite{yeong16} proposed the use of LSTM to make the system focus on different frames of videos, but it was unable to represent intervals.

The main contribution of this paper is the introduction of the temporal attention filters that allow their end-to-end training together with underlying CNN architectures. We illustrate that our temporal attention filters can be learned to focus on different temporal aspects of videos (i.e., intervals with different temporal resolutions), and experimentally confirm that such learning benefits the activity recognition. The main difference between our approach and previous temporal structure learning methods for activity recognition (e.g., \cite{niebles10,ryoo13}) is that our temporal filters are designed to be fully differentiable, which allows their joint learning and testing with modern CNN architectures.

To our knowledge, this paper is the first paper to enable learning of latent temporal sub-events in an end-to-end fashion using CNN architectures for activity recognition.



\section{Recognition approach}
\label{sec:approach}

We design our model as a set of temporal filters, each corresponding to a particular sub-event, placed on top of per-frame (or per-segment) CNN architectures (Figure \ref{fig:pipeline}). The idea is to train this fully differential model in an end-to-end fashion, jointly learning latent sub-events composing each activity, underlying CNN parameters, and the activity classifier.

In this section, we introduce the concept of \emph{temporal attention filters}, which extends the spatial attention filter \cite{attention15} originally designed for digit detection and digit/object synthesis. Next, we present how our proposed model takes advantage of temporal attention filters to learn latent sub-events. The approach learns/mines temporal sub-events optimized for the classification without their ground truth annotations, such as `stretching an arm' in the activity `punching'.

\subsection{Temporal attention filters}

\begin{figure}[b]
\begin{center}
   \includegraphics[width=1.0\columnwidth]{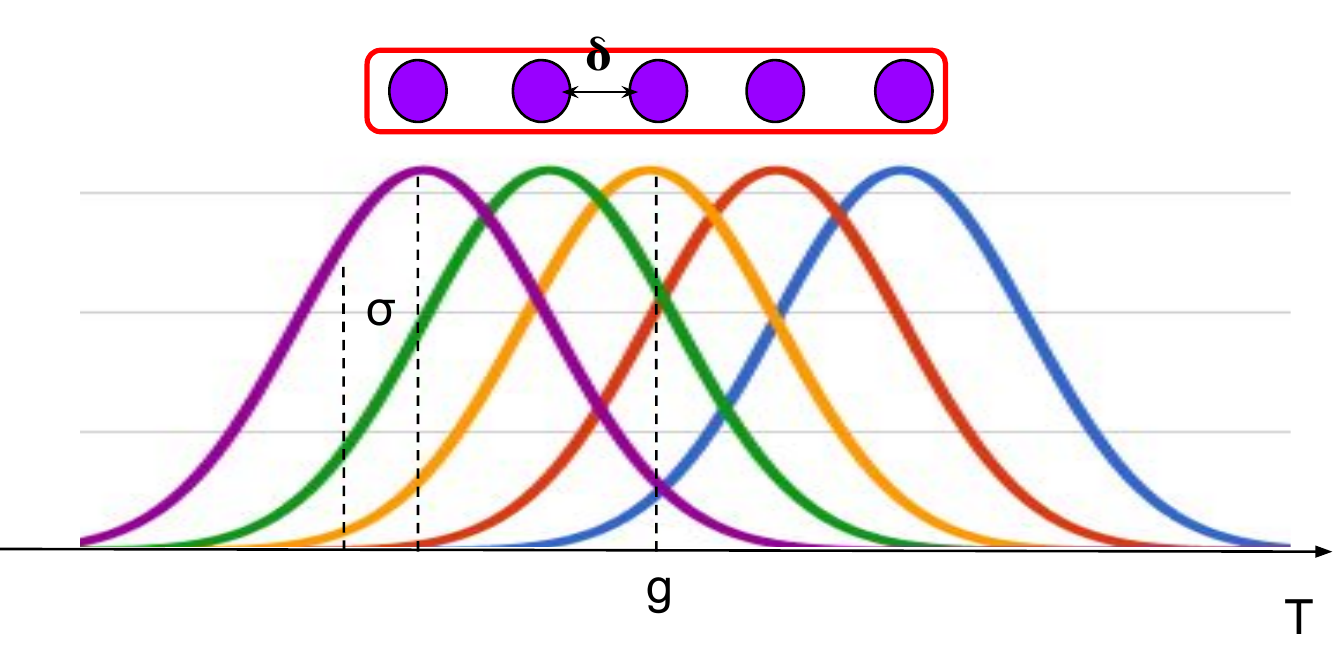}
\end{center}
   \caption{An illustration of our temporal attention filter. The filter is differentiable and represented with three parameters.}
\label{fig:filter}		
\end{figure}

Each temporal attention filter learns three parameters: a center $g$, a stride $\delta$ and a width $\sigma$. These parameters determine where the filter is placed and the size of the interval focused on. A filter consists of $N$ Gaussian filters separated by a stride of $\delta$ frames. The goal of this model is to learn where in the video the most useful features appear. Because the videos are of variable length, $\delta$ and $g$ are relative to the length of the video. Based on the attention model presented in~\cite{attention15}, we use the following equations to obtain the mean of the Gaussian filters:
\begin{equation} \label{eq:mrf_match}
\begin{split}
g_n &= 0.5\cdot T \cdot (\widetilde{g}_n + 1) \\
\delta_n &= \frac{T}{N-1} \widetilde{\delta}_n \\
\mu_n^i &= g_n + (i - 0.5N +0.5)\delta_n \\
\end{split}
\end{equation}

Using $\mu$ and $\sigma$, the $N$ Gaussian filters are defined by:
\begin{equation} \label{eq:mrf_match}
\begin{split}
F_m[i,t] &= \frac{1}{Z_{m}} \exp(-\frac{(t-\mu_m^i)^2}{2\sigma_m^2}) \\
& i\in\{0,1,\ldots,N-1\},~ t\in\{0,1,\ldots,T-1\} \\
\end{split}
\end{equation}
where $Z_m$ is a normalization constant. Figure \ref{fig:filter} shows an illustration of our temporal attention filter.

If each frame has $D$-dimensional features, the filters, $F$, are applied to each dimension, taking the input of size $T\times D$ to $N\times D$ where $T$ is the number of frames in the video. That is, each temporal filter $F$ generates a $N\times D$-dimensional vector as an output for any video, which can be passed to a neural network for the classification. Since this model is fully differentiable, all parameters can be learned using gradient descent.

Let $f_{m}[i,d]$ be the output of our $m$th temporal attention filter, given the $T\times D$ dimensional input $x$. Each $f_{m}[i,d]$ describes the response from the $i$th Gaussian filter on the $d$th elements of the input vectors. Then,
\begin{equation} \label{eq:read}
\begin{split}
f_{m}[i,d] &= F_m[i,:] \cdot x[:,d] = \sum_{t=0}^{T-1} F_m[i,t] \cdot x[t,d] \\
& i\in\{0,1,\ldots,N-1\},~ d\in\{0,1,\ldots,D-1\} \\
\end{split}
\end{equation}
where $x[:,d]$ is the $T$-dimensional vector corresponding to the sequence of the $d$th element values in the underlying CNN feature vectors.

\begin{figure}[!tbp]
\begin{center}
   \includegraphics[width=1.0\columnwidth]{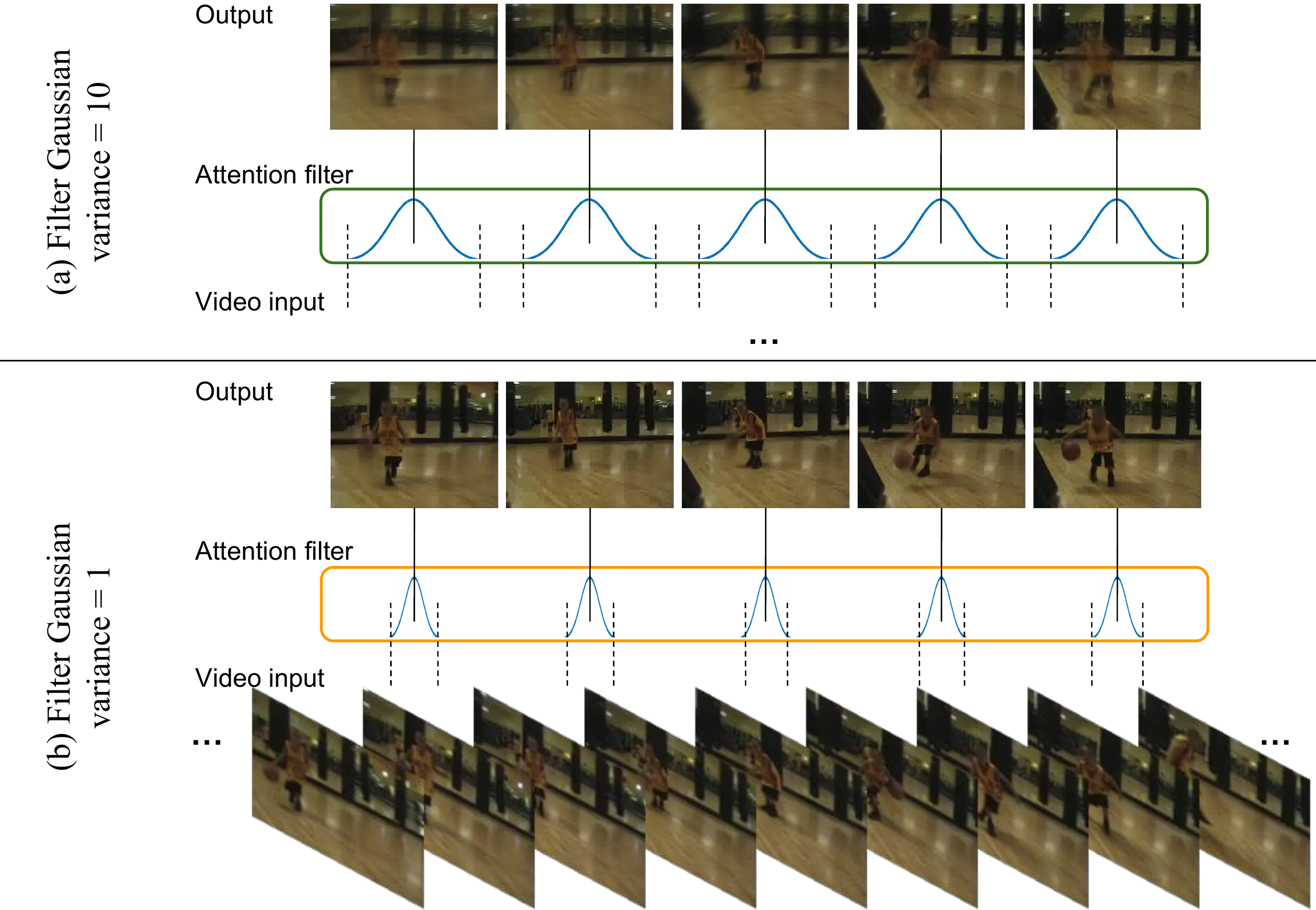}
\end{center}
   \caption{Examples illustrating how our temporal attention filters work once learned. They are shown with two different Gaussian filter variances. For this visualization, raw video frames are used as inputs directly. In our actual implementation, the input to temporal filters are not raw video frames but CNN feature responses obtained from frames.}
\label{fig:filter_example}		
\end{figure}

Figure \ref{fig:filter_example} shows how each temporal attention filter is able to capture features from the corresponding sub-interval of the provided video with different temporal resolutions.

\subsection{Recognition with temporal attention filters}
\label{subsec:recognition}

As described in Figure \ref{fig:pipeline}, we take advantage of multiple different temporal attention filters by placing them on top of a sequence of per-frame (or per-segment) CNN models. As a result, our model is able to focus on different sub-intervals of video inputs with different temporal resolutions. Outputs of each temporal filters are concatenated and are connected to fully connected layers performing activity classification. Each of our filters learns a latent sub-event, and concatenating the results allows the later FC-layers to look at the features for each sub-event and classify the activity based on them. If we denote the per-frame feature size as $D$ and we have $M$ number of temporal attention filters, each temporal filter generates the output of size $N\times D$, resulting the total dimensionality to be $M\times N\times D$. We used 2 fully connected layers (i.e., one hidden layer and one soft-max layer) for the classification. 

Because of the property that our temporal filters are designed to be differentiable, we are able to backpropagate the errors through temporal attention filters reaching the underlying per-frame convolutional layers. This makes end-to-end training of the proposed model possible with video training data. Per-frame CNNs were assumed to share all parameters.


\subsection{Recurrent neural networks with temporal filters}
\label{subsec:lstm}

Although the model presented in the previous subsection allows us to learn temporal attention filters from the training data, it was assumed that the temporal filters are static once learned and are shared across all videos. However, such assumption that relative locations of sub-events are exactly identical across all activity videos can be dangerous. A particular sub-event of the activity (e.g., a person stretching an arm in the case of `shake hands') may occur earlier or faster in one video than those in the other videos, due to human action style variations. In such cases, using static temporal filters will fail to capture exact sub-events. Rather, the recognition system must learn how to dynamically and adaptively adjust locations of temporal filters depending on the video content.

\begin{figure}[!tbp]
\begin{center}
   \includegraphics[width=1.0\columnwidth]{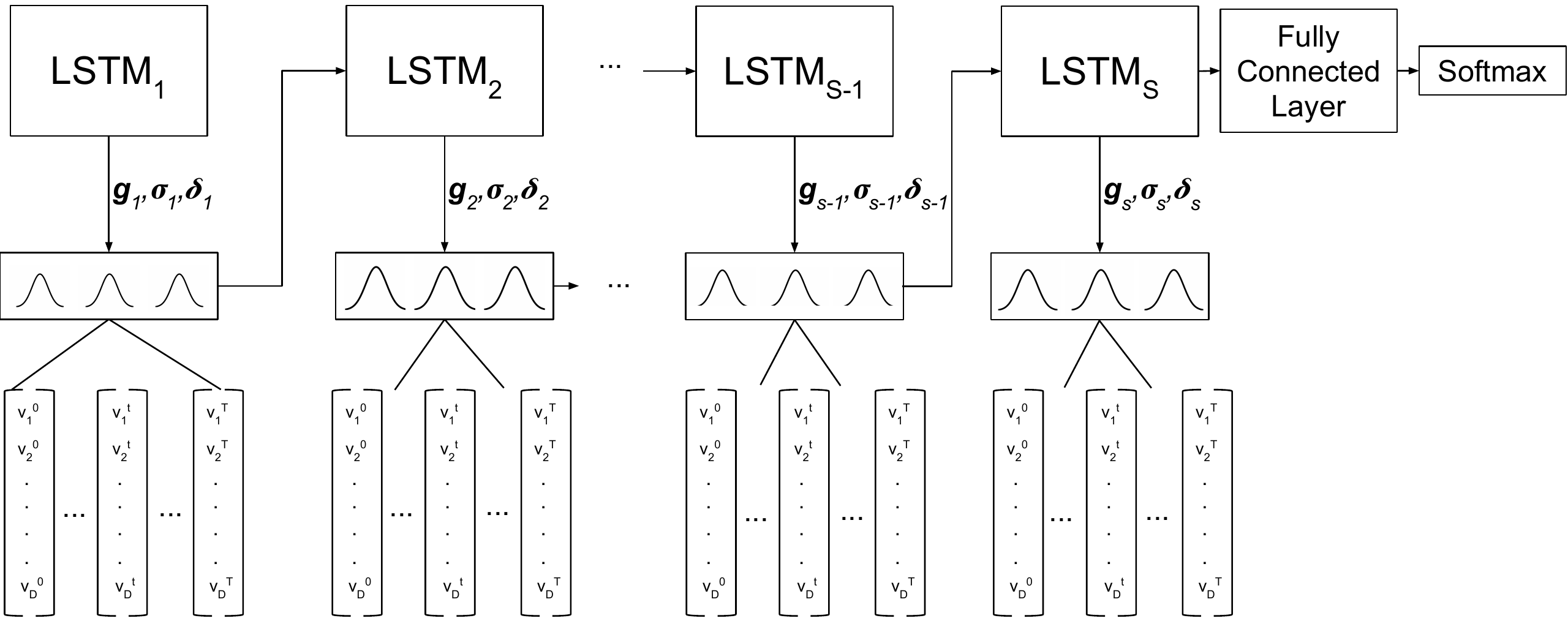}
\end{center}
   \caption{An illustration of our temporal attention filter using an LSTM. The LSTM provides the parameters for the filters at each time step. The filters provide input to the next iteration allowing the LSTM to adjust the location of the filters. After $S$ iterations, a fully-connected layer and softmax layer classify the video. }
\label{fig:lstm_model}		
\end{figure}


Thus, in this subsection, we propose an alternative approach of using a recurrent neural network, the long short-term memory (LSTM). Figure~\ref{fig:lstm_model} describes our overall LSTM architecture. At each iteration, our LSTM takes the entire video frames as an input and applies per-frame CNNs identical to the previous subsection. Next, instead of using the learned static temporal filters, previous LSTM outputs are used to decided the temporal attention filter parameters in an adaptive fashion. Our approach learns weights that models how previous LSTM iteration outputs (i.e., the abstraction of video information in the previous round) can lead to the better temporal filters in the next iteration

More specifically, our attention filter parameters become the function of previous iteration LSTM outputs:
\begin{equation} \label{eq:mrf_match}
\begin{split}
(g_t, \delta_t, \sigma_t) = W_n(h_{h-1}) = \sum_i w_i \cdot h_{t-1}(i)
\end{split}
\end{equation}
where $W_n$ is the function we need to learn modeled as a weighted sum, $h_{t-1}$ is the LSTM hidden state vector at iteration $t-1$ and $h_{t-1}(i)$ are its elements. These weights are initialized such that the initial iteration places $g$ at the center of the video, and $\delta$ spans the entire duration, allowing the LSTM to get input from the entire sequence of frames. Because of the nature that our temporal attention filters are differentiable, we learn the function $W_n$ through the backpropagation.

Notice that our usage of LSTMs is different from previous approaches \cite{google15} using LSTMs to capture sequential per-frame feature changes. In our case, the goal of each LSTM iteration is to adjust the temporal filter locations to match the video input.


\section{Experiments}

In order to evaluate the effectiveness of our proposed latent sub-event learning approach with fully differentiable temporal attention filters, we conducted a set of experiments comparing our approach using temporal filters against the previous conventional approaches without them while making the approaches use the same features and classifiers. These approaches were evaluated with multiple different features and multiple different datasets.

{\flushleft\textbf{Features:} We extracted VGG features \cite{vgg}, INRIA's improved trajectory features (ITF) \cite{wang13}, C3D features \cite{c3d}, and TDD features \cite{tdd15} which were used as inputs to our temporal filter model. VGG is an image-based convolutional neural network originally designed for object classification tasks. ITF and TDD are the state-of-the-art trajectory-based local video features, each taking advantage of HOG/HOF or CNN feature maps observed around of trajectories. We used the source codes of all these features provided by the authors of the corresponding papers. For TDD, we used the single-scale version of the TDD feature, since it showed the better performance. We used 3-frame short video segments as our unit observations. In the case of VGG features, we applied its CNN architecture to one image frame in every 3 frames, obtaining 4K-dimensional vectors from the final fully connected layer. In the case of trajectory features, we considered the trajectories ending within the 3-frame segment as inputs corresponding to the segment and took advantage of their Fisher vector representation.}


{\flushleft\textbf{Datasets:} We conducted experiments with two different public video dataset: DogCentric activity dataset \cite{ryoo14dog} and HMDB dataset \cite{kuehne11}. The DogCentric dataset is a first-person video dataset, and it was chosen because that the previous pooled time series (PoT) representation \cite{ryoo15}, which illustrated potential that considering multiple sub-intervals of videos benefit the CNN-based recognition, achieved the state-of-the-art performance on it.  The first-person videos in this dataset display a significant amount of ego-motion (i.e., camera motion) of the camera wearer, and it is an extremely challenging dataset. HMDB was chosen due to its popularity. Our experiments were conducted by following each dataset's standard evaluation setting. Both the datasets are designed for multiclass activity video classification.}

{\flushleft\textbf{Implementation and baseline classifiers:} As mentioned above, we confirm the advantage of our approach with four different types of underlying feature models: VGG, ITF, C3D, and TDD. Our temporal filters were applied on top of them. In all our experiments, we used 2 hidden layers on top of our (multiple) temporal filters: one of size 2048 and one of size 10 and softmax classification.}

As the basic baselines, we tested (1) max-pooling, (2) sum-pooling, and (3) mean-pooling across the time dimension, resulting in features of size $1\times D$. These $1\times D$ features were fed to the hidden layers for the classification, which was a standard practice as described in \cite{jain14,karpathy14,simonyan14,google15,c3d}. In addition, in order to explicitly confirm the power of our approach of `learning' temporal attention filters with different location/duration/resolution, we also implemented the baselines of using fixed-predetermined temporal filters (i.e., without learning). The main idea was to make the systems take advantage of temporal filters identical to ours while disabling their learning capability. This makes the baselines behave very similar to previous pooled time series method \cite{ryoo15}. We tested (4) an approach of using a single fixed filter (i.e., the level 1 temporal pyramid), which essentially computes a weighted sum of the whole video. We then used (5) a temporal pyramid of level 4, having 15 filters: 1 viewing the whole video, 2 viewing half the video, 4 viewing a forth, and 8 viewing an eighth, giving a vector of size $15\times N\times D$.

We implemented our CNN-based recognition architecture with learnable temporal attention filters as described in this paper. First, we implemented our approach of learning static temporal filters. We tested our model's ability to learn filters by using 15 filters (i.e., $M=15$) with $N=1$ or $N=3$. Finally, we modified the model to use a LSTM to dynamically choose where to look. At each step, the temporal model took the hidden state, $h$ from the LSTM as input, and did a linear transformation with learned weights and bias $W\cdot h + b$ to obtain $g, \delta$ and $\sigma$. We then used the same equations as above to create the filters. The LSTM ran for 4 steps and either had 1 or 3 filters (i.e., $M=1$ or $3$) with $N=5$.

{\flushleft\textbf{Training the network:} To increase training data, we apply random cropping on each video frame, and randomly skip several video frames at beginning.  We use log-scale of stride and variance to ensure positivity as~\cite{attention15}.
We initialize each filter bank parameters $(\widetilde{g}_m, \log\widetilde{\delta}_m, \log\sigma_m^2)$ with normal distribution for $\widetilde{g}_m$  and 0 for $\log\widetilde{\delta}_m$ and $\log\sigma_m^2$.}

For all experiments, the first fully-connected layer had 4096 nodes and used a ReLU activation function. The second had either 10 nodes (for DogCentric) or 51 nodes (for HMDB) and used soft-max. The network was trained for 10000 iterations with a batch size of 100 and stochastic gradient descent with momentum set to 0.9. 

\subsection{DogCentric dataset}
The DogCentric dataset consists of 209 videos (102 training and 107 testing videos) and 10 classes. As mentioned above, it is a very challenging dataset with severe camera motion.

\begin{figure}[!tbp]
\begin{center}
   \includegraphics[width=1.0\columnwidth]{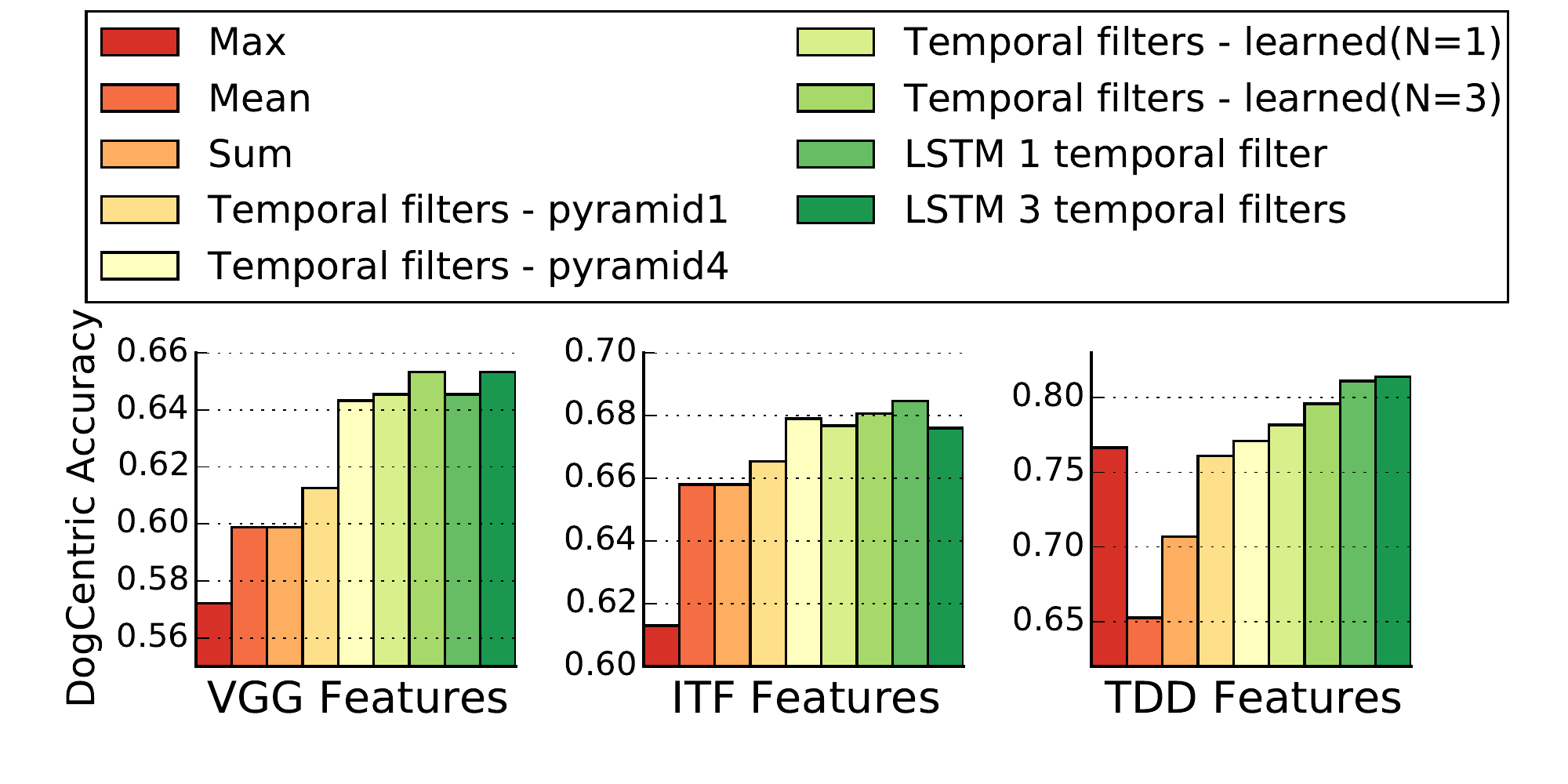}
\end{center}
\vspace{-20pt}
   \caption{Classification accuracy of baselines and our proposed approach using VGG, ITF and VGG features.}
\label{fig:dog_centric_plot}		
\end{figure}

All of the baselines described above as well as 4 different versions of our approach (2 with static filter learning and 2 with LSTMs) were compared while using 3 different types of underlying features. Figure \ref{fig:dog_centric_plot} shows the results. We are able to clearly observe that the consideration of multiple sub-intervals improves the recognition performances. The performance increased by using predetermined temporal pyramid, and our proposed approach of learning temporal filters were able to further improve the performance. Additionally, using the LSTM to dynamically choose sub-event locations gives the best performance.

The overall difference between the conventional approach of max/sum/mean pooling and our learned temporal attention filters are around 5\% in VGG, 1\% in ITF, and 4$\sim$5\% in TDD. We believe our approach was more effective with VGG since it's a pure CNN-based feature allowing our differentiable filters to better cope with them. ITF is a completely hand-crafted feature and TDD is a partially hand-crafted feature.

Table \ref{table:dog-compare} compares the performances of our approach with the previously reported state-of-the-arts. The base features we used for this table was TDD. By taking advantage of temporal filter learning and also LSTMs to dynamically adjust the filters matching sub-events, we were able to outperform the state-of-the-arts.

\begin{table}
    \small
	\caption{Recognition performances of our approach on the DogCentric dataset, compared against previously reported results of state-of-the-art approaches.}
	\label{table:dog-compare}

	\center
	\setlength\extrarowheight{0pt}

		\begin{tabular}	{c|c}
			\hline 	Approach & Accuracy \tabularnewline
			\hline 	
			        VGG \cite{vgg}	& 	59.9 \%  \tabularnewline
			        Iwashita et al. \citeyear{ryoo14dog}    &   60.5 \%  \tabularnewline
			        ITF \cite{wang13}	& 	67.7 \%  \tabularnewline
			        ITF + CNN \cite{jain14}	& 	69.2 \%  \tabularnewline
			        PoT \cite{ryoo15}    &   73.0 \%  \tabularnewline
			        TDD \cite{tdd15} &   76.6 \%  \tabularnewline
			\hline
					Ours (temporal filters)	&   79.6 \%  \tabularnewline
					Ours (temporal filters + LSTM)  &  \bf{81.4} \% \tabularnewline
			\hline
		\end{tabular}

\end{table}

\subsection{HMDB}

HMDB is a relatively large-scale video dataset with 51 activity classes and more than 5000 videos.

In our first experiment, we made our model to learn general temporal filters to be \emph{shared} by all 51 activity classes, testing whether our approach can learn globally optimal filters. Table \ref{table:hmdb_results} shows the results of our approach with the temporal attention filters compared against the approaches without them. We tested this with CNN-based features: VGG, C3D, and TDD. We are able to observe that our approach of learning latent sub-events using temporal attention filters clearly benefits the classification in all cases. 

\begin{figure*}[!tbp]
    \begin{center}
        \includegraphics[width=0.89\linewidth]{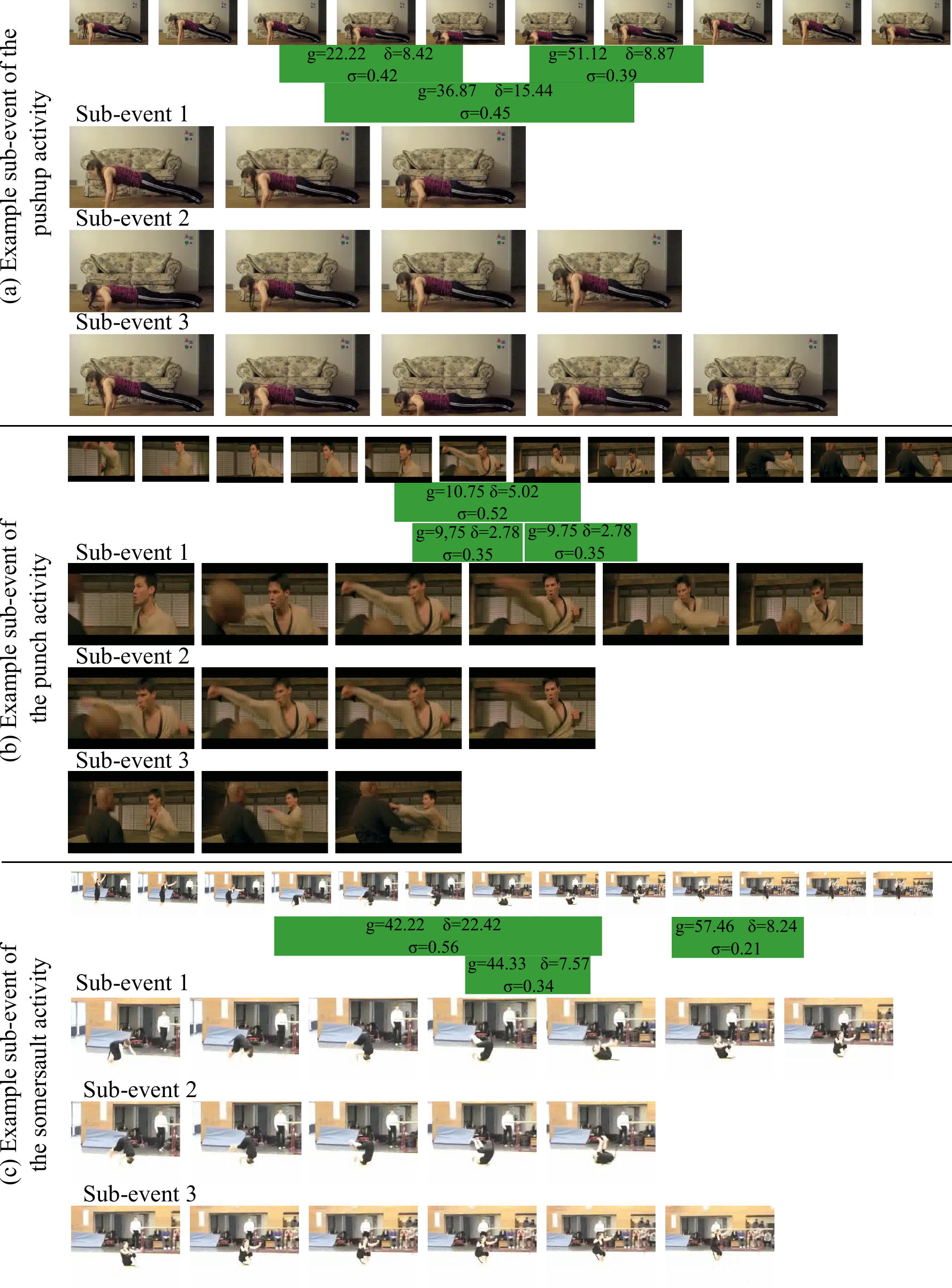}
    \end{center}
    \vspace{-20pt}
   \caption{Example frames of the learned latent sub-events. The top row of each sub-figure (a-c) shows an example frame sequence, and the green boxes below shows the locations of the temporal filters learned to capture (latent) sub-events. Actual frames corresponding to each sub-event are also illustrated. We are able to observe that semantic sub-events such as (a) `doing down' in the pushup activity, (b) `stretch' in the punch activity and (c) `roll' and `stand up' in the somersault activity are captured.}
\label{fig:subevent_example}		
\end{figure*}

Finally, instead of making our model to learn general filters, we made our models to learn 1-vs-all binary classifier for each activity. This enables each activity classification model to learn class-specific (latent) sub-events tailored for the activity, allowing the model to fully take advantage of our filter learning method. These 1-vs-all results were combined for the final 51-class classification. We were able to get an accuracy of 68.4\% (Table \ref{table:hmdb-compare}). This is significant considering that the base feature performance (i.e., TDD) is 57\% with max/mean pooling. The setting was $N$ = 2 and $M$ = 3. The performance increase gained by learning such latent sub-events per activity was significant and consistent: it was 9\% increase over fixed temporal pyramid filters with C3D and 10\% increase over temporal pyramid with TDD. There are few existing approaches performing comparable to our approach by fusing multiple features (e.g., 69.2 of 
\cite{feichtenhofer2016} + ITF and 67.2 of \cite{varol16} + ITF), but our approach using one feature performed superior to them in their original form. Figure \ref{fig:subevent_example} shows examples of the learned sub-events.


\begin{table}[]
\small
\caption{A table comparing the performance of the approaches using the HMDB Dataset. In this experiment, we made the model to learn general temporal filters \emph{to be shared} across all 51 activities.}
\label{table:hmdb_results}

\center
\setlength\extrarowheight{0pt}
\begin{tabular}{c|c|c|c}\hline
Method       &  VGG & C3D & TDD \\\hline
\textbf{Baseline}       \\\hline
Max Pooling  &  37.77 \% &  48.45 \%  &  57.07 \%  \\
Sum Pooling  &  37.00 \% &  48.58 \%  &  55.77 \%   \\
Mean Pooling &  37.73 \% &  49.30 \%  &  57.17 \%  \\\hline
\textbf{Fixed Temporal Filters}       \\\hline
Pyramid 4  &  41.56 \%   & 49.69 \%  &  58.87 \%\\\hline
\textbf{Learned Temporal Filters}       \\\hline
$N=1$           &  41.23 \%  & 50.35 \%  &  58.87 \% \\
$N=3$           &  42.50 \%  & 50.00 \%  &  59.03 \%  \\
LSTM 1 filter   &  42.72 \%  & 51.20 \%  &  58.04 \%  \\
LSTM 3 filters  &  43.03 \%  & 49.81 \%  &  58.93 \%  \\\hline
\end{tabular}
\end{table}

\begin{table}
    \small
	\caption{Final recognition performances of our approach with \emph{per-activity} temporal filters, tested with HMDB. Results are compared against the state-of-the-arts.}
	\label{table:hmdb-compare}

	\center
	\setlength\extrarowheight{0pt}
		\begin{tabular}	{c|c}
			\hline 	Approach & Accuracy \tabularnewline
			\hline 	
			        ITF \cite{wang13}	& 	57.2 \%  \tabularnewline
			        2-stream CNN \cite{simonyan14}	& 	59.4 \%  \tabularnewline
			        TDD \cite{tdd15} &   63.2 \%  \tabularnewline
			        LTC \cite{varol16} &   64.8 \%  \tabularnewline
			        S+T \cite{feichtenhofer2016} &   65.4 \%  \tabularnewline
			        \hline
			        Max pooling - C3D   &  48.5 \%   \tabularnewline
			        Temporal pyramid - C3D    &  49.7 \%   \tabularnewline
			        Ours (temporal filters) - C3D   &  57.7 \%   \tabularnewline
			        \hline
			        Max pooling - TDD   &  57.1 \%   \tabularnewline
			        Temporal pyramid - TDD    &   58.9 \%   \tabularnewline
					Ours (temporal filters) - TDD	&   \textbf{68.4} \%  \tabularnewline
			\hline
		\end{tabular}
\end{table}

\section{Conclusion}

We present a new activity recognition approach using \emph{temporal attention filters}. Our approach enables end-to-end learning of the classifier jointly with latent sub-events and underlying CNN architectures. An approach to adaptively update the temporal filter location/duration/resolution using multiple recurrent LSTM iterations was also proposed, and its potential was experimentally confirmed. The concept of learning latent sub-events using temporal attention clearly benefited the recognition, particularly compared to the baselines using the same features without such learning.



{\flushleft\textbf{Discussions:} Unfortunately, we were unable to replicate the TDD's reported recognition performance of 63.2 \% with the code provided by the authors. We only obtained 57\%. This probably is due to the difference in detailed parameter settings and engineering tricks. If we can replicate the performance of TDD reported in its paper, we would be able to further increase our method's performance using it as a base.}

\section*{Acknowledgement}

This work was supported by the ICT R\&D program of MSIP/IITP [16ZC1310, Development of XD Media Solution for invigoration of Realistic Media Industry]. This work was also supported in part by the Army Research Laboratory under Cooperative Agreement Number W911NF-10-2-0016.


\bibliographystyle{aaai}
\bibliography{tempbib}

\end{document}